\documentclass[conference]{IEEEtran}
\IEEEoverridecommandlockouts
\usepackage{cite}
\usepackage{amsmath,amssymb,amsfonts}
\usepackage{algorithmic}
\usepackage{graphicx}
\usepackage{textcomp}
\usepackage{caption}
\usepackage{subcaption}
\usepackage{xcolor}
\usepackage{svg}
\usepackage{amsmath}
\usepackage[symbol]{footmisc}

\makeatletter
\def\@fnsymbol#1{\ensuremath{\ifcase#1\or \dagger\or \ddagger\or
   \mathsection\or \mathparagraph\or \|\or **\or \dagger\dagger
   \or \ddagger\ddagger \else\@ctrerr\fi}}

\usepackage{hyperref}

\def\BibTeX{{\rm B\kern-.05em{\sc i\kern-.025em b}\kern-.08em
    T\kern-.1667em\lower.7ex\hbox{E}\kern-.125emX}}

\begin{document}

\title{Merging Tasks for Video Panoptic Segmentation}

\author{
    Jake Rap \\
    j.e.w.rap@student.tue.nl
  \and
    Supervisor\\
    Panagiotis Meletis\\
    p.c.meletis@tue.nl
}

\maketitle
\thispagestyle{plain}
\pagestyle{plain}

\begin{abstract}
In this paper, the task of video panoptic segmentation is studied and two different methods to solve the task will be proposed. Video panoptic segmentation (VPS) is a recently introduced computer vision task that requires classifying and tracking every pixel in a given video. The nature of this task makes the cost of annotating datasets for it prohibiting. To understand video panoptic segmentation, first,  earlier introduced constituent tasks that focus on semantics and tracking separately will be researched. Thereafter, two data-driven approaches which do not require training on a tailored VPS dataset will be selected to solve it. The first approach will show how a model for video panoptic segmentation can be built by heuristically fusing the outputs of a pre-trained semantic segmentation model and a pre-trained multi-object tracking model. This can be desired if one wants to easily extend the capabilities of either model. The second approach will counter some of the shortcomings of the first approach by building on top of a shared neural network backbone with task-specific heads. This network is designed for panoptic segmentation and will be extended by a mask propagation module to link instance masks across time, yielding the video panoptic segmentation format. \\




\end{abstract}


\section{Introduction}\label{ch: Introduction}


Understanding dynamic scenes is of primary importance for mobile autonomous systems. For an autonomous car or robot to navigate through complex environments, its perception system has to tackle the challenging task of extracting useful information from what it perceives. This task is holistic in nature and can thus be split up into multiple constituent tasks that in themselves might be easier to solve. A perception system can, for example, be designed to locate, classify and track objects or their environment. Besides being able to successfully solve those separate tasks, it continues to be challenging to find ways to combine their results such that a more complex emergent task can be solved.


Since the advent of deep learning, the area of computer vision has seen tremendous improvements and increasing interest in algorithms based on deep neural networks. These data-driven learning approaches nowadays outperform algorithms based on classical computer vision and heuristics alone.\\
On the way to designing a truly comprehensive perception system, multiple different constituent tasks focused on semantics and tracking have been defined that cover parts of the overall holistic task. Even though the state of the art in these constituent tasks keeps improving, the pace at which their performance grows is saturating \cite{HAI2021}. This leads more and more of current scene understanding research to aim at combining these tasks, in an effort to keep pushing the limits and work towards designing an all-encompassing perception system.


The recently defined task of video panoptic segmentation (VPS) \cite{Kim2020} proposes an interesting view on how to tackle building an end-to-end model for holistic scene understanding. Indeed it shows how a more complex task can emerge from multiple constituent tasks. Some current limitations for VPS include the need for large densely labeled datasets and the tracking performance drop for longer sequences.\\

\begin{figure}[t]
    \begin{subfigure}{\linewidth}
        \includegraphics[width=\linewidth]{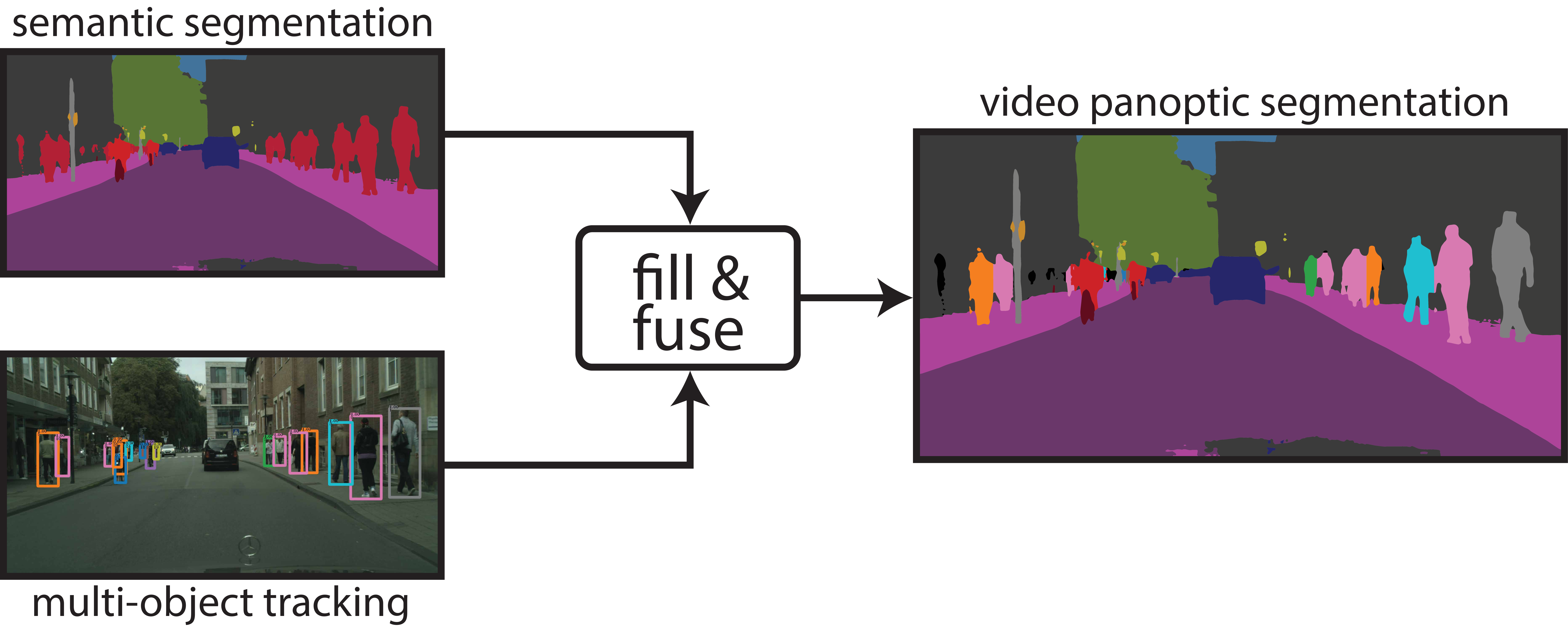}
        \caption{Using the designed ``Fill \& Fuse" module, the outputs of the tasks semantic segmentation and multi-object tracking can be converted to the video panoptic segmentation task format.\\}
    \end{subfigure}

    \begin{subfigure}{\linewidth}
        \includegraphics[width=\linewidth]{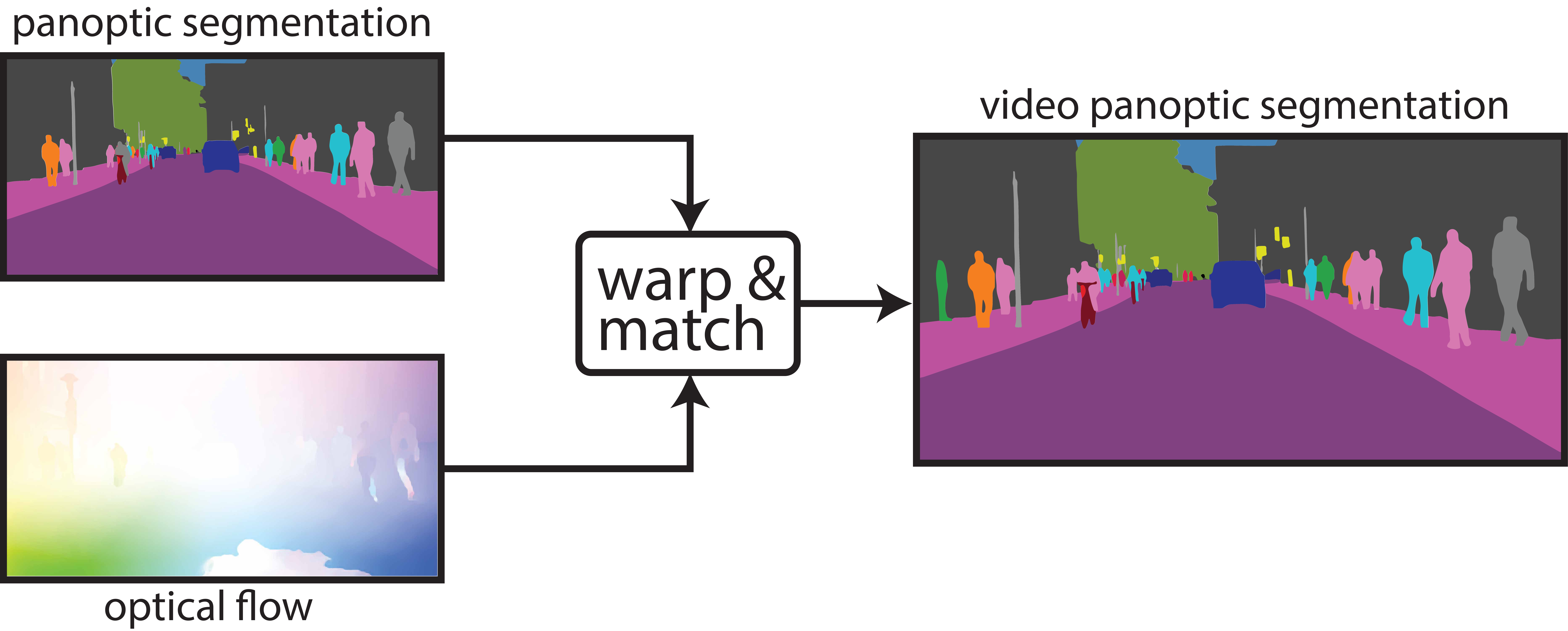}
        \caption{Using the designed ``Warp \& Match" module, the outputs of the task (image) panoptic segmentation can be made time consistent by utilizing an optical flow map.}
    \end{subfigure}

    \caption{An overview of the methods used in this work. For both approaches, a module has been designed that can transform two specific computer vision task outputs into the format of the video panoptic segmentation task.}
    \label{fig:eye_catcher}
\end{figure}

The primary aim of this paper is to investigate the aspects of video panoptic segmentation in order to select an algorithm for solving this task. This will be done by conducting a literature study on the task itself and its constituent tasks, as well as by proposing two distinctive modules, see Fig. \ref{fig:eye_catcher}, that can be added on top of other existing networks to extend them to the video panoptic segmentation task without the need for retraining or VPS-compatible datasets.
\newpage
In summary, the main contributions of this paper are:

\begin{itemize}
  \item We propose a VPS solution that uses the our ``Fill \& Fuse" module, which combines the outputs of a semantic segmentation network and multi-object tracking network and transform them into the video panoptic segmentation output format.
  \item We propose a VPS solution that uses our ``Warp \& Match" module, which utilizes an optical flow network to make the outputs of a panoptic segmentation network time consistent.
\end{itemize}

\section{Related Work}\label{ch: related_work}



\subsection{Panoptic Segmentation}
A major step towards real-world vision systems was introduced by \cite{Kirillov2019}, with the task proposition and definition of panoptic segmentation. Panoptic segmentation is a unification of two earlier introduced tasks, namely semantic segmentation \cite{farabet2012scene} and instance segmentation \cite{gupta2014learning}. The goal of semantic segmentation is to classify each pixel in an image and assign it a class label $c$. Whereas, the goal of instance segmentation is to find every countable object in an image and delineate it with a segmentation mask. By bringing these two tasks together we get panoptic segmentation.
In the area of segmentation and object detection there is an important distinction between two types of classes. On the one hand, we have classes to which objects belong that are countable (e.g. car, pedestrian, cyclist), these classes are commonly referred to as $thing$ classes and on the other hand, we have classes to which uncountable objects belong (e.g. grass, road, sky), these are referred to as $stuff$ classes. Instance aware segmentation of stuff, as well as thing classes yields panoptic segmentation, which can be defined as labeling every pixel of an image with a class label and additionally labeling every pixel belonging to a $thing$ class with an instance id. Pixels belonging to $stuff$ classes do not need an instance id.

\begin{figure*}
\includegraphics[width=\textwidth]{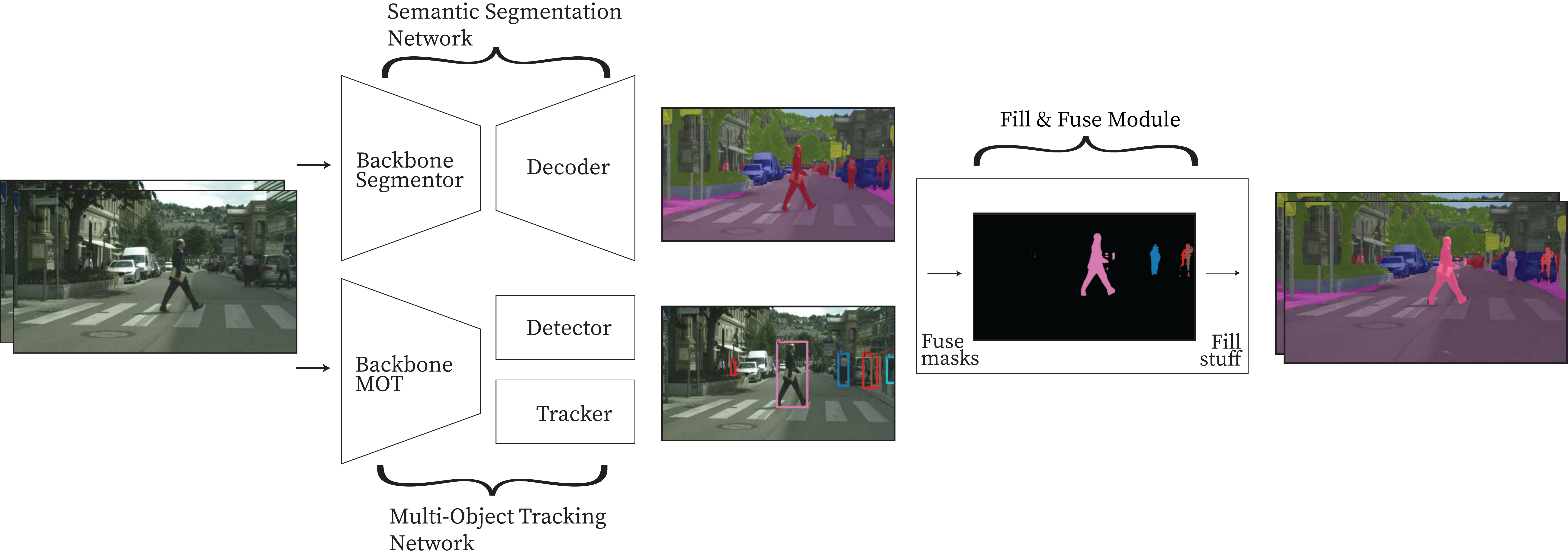}
\caption{First baseline for a video panoptic segmentation model. It is comprised out of a semantic segmentation network and a multi-object tracking network, which outputs will be fused together in the fill \& fuse module to create the VPS format.}
\label{fig: model1}
\end{figure*}

\subsection{Multi-Object Tracking}
By building on top of frameworks for object detection \cite{DBLP:journals/corr/RenHG015}, the authors of \cite{shuai2020multiobject} create a baseline for multi-object tracking. For this task, it is required to detect object instances (i.e. $things$) in a frame of a video and to track these instances as they move through consecutive video frames. As opposed to an algorithm for a task like panoptic segmentation, an algorithm for multi-object tracking needs a notion of dynamics and the time dimension. The detection and tracking of instances for multi-object tracking tasks is done by using bounding boxes that encapsulate the objects in a video. Bounding boxes however are a coarse way to describe the location of an object as they can overlap and do not represent the boundaries of objects as precisely as a segmentation mask.\\
The multi-object tracking task has seen enormous research interest the last couple of years, especially since the introduction of the MOTChallenge benchmark \cite{Dendorfer2021}, which primarily focuses on the online tracking of pedestrians in videos. Benchmarks like these have been crucial in pushing the performance of computer vision algorithms. They provide standardized metrics and large data sets which are important for the development of deep neural networks.

\subsection{Towards Video Panoptic Segmentation}
As bounding box level tracking is saturating we must move to pixel-level tracking in order to keep pushing performance \cite{Voigtlaender2019}. Multi-object tracking and segmentation \cite{Voigtlaender2019} achieves pixel-level tracking by segmenting and tracking $thing$ classes and is thus a unification of instance segmentation and multi-object tracking. This trend continued with \cite{Hurtado2020} and the definition of multi-object panoptic tracking, which is a unification of multi-object tracking and panoptic segmentation, a task with a very similar format and goal as video panoptic segmentation. On the other hand of the spectrum, the real power and relevance of using panoptic segmentation is if this task can be used in the video or time dimension. Therefore \cite{Kim2020} defines video panoptic segmentation. In video panoptic segmentation we want to output a class label and an instance ID per pixel, per frame, and want these predictions to be consistent throughout all video frames. The prime challenge in VPS is that we want our panoptic segmentations of frames to be consistent across time. The object that in one frame might have the instance ID of $car1$ still has to have the same instance ID in any consecutive frame. By having time-consistent instance IDs we can track objects in a video frame sequence. \\

\subsection{Evaluation Metrics}
An important aspect to stimulate research for a specific task is to have a common and accepted way to assess the performance of an algorithm designed for it. The formal definition of video panoptic segmentation by \cite{Kim2020} is also accompanied by the evaluation metric, video panoptic quality (VPQ). VPQ is an extension of the metric panoptic quality (PQ) \cite{Kirillov2019} in the video dimension. While designing a metric for a task that is build on top of other tasks, one can draw inspiration from formerly introduced benchmarked metrics. This, and the imbalance in VPQ led \cite{Weber2021} to investigate metrics that came originally from the tracking domain i.e. metrics designed for multi-object tracking. The downsides and upsides of metrics like MOTA \cite{Bernardin2008}, HOTA \cite{Luiten2021},  VPQ \cite{Kim2020}, and (s)PTQ \cite{Hurtado2020} were all investigated and a new metric and was introduced. The segmentation and tracking quality (STQ) \cite{Weber2021} combats most problems earlier metrics had and fits the vision of the comprehensive task of video panoptic segmentation.

\subsection{Datasets}
Another important part of working with data-driven approaches for computer vision is the datasets needed to train models. Due to the complexity of video panoptic segmentation, the datasets used to train a model for it require many finely labeled images. Labeling these images takes expensive manual work. Therefore, the datasets used to train models for tasks like these are not fully annotated. The Cityscapes dataset \cite{cordts2016cityscapes} contains 25,000 images of which 5,000 finely labeled for panoptic segmentation. These images belong to 30-frame video snippets, each with a duration of 1.8s (17Hz). Even though this is already a lot of data, the labels for this dataset are too spread out over time, since only one frame (the $20^{th}$) of each video snippet is labeled. With the emergence of video panoptic segmentation, the time-consistency plays a crucial role. Having only one frame annotated per snippet did not suffice anymore. Cityscapes-VPS \cite{Kim2020} therefore iterated over the original dataset by finely labeling 5 more frames every video snippet in the validation set. Even though still not every frame is labeled, the dataset can still successfully be used to train models for the time-consistent video version of panoptic segmentation.



\section{Methods}\label{ch: method}
To achieve video panoptic segmentation, two different baselines have been designed and studied in this work. The first baseline fuses readily available neural networks designed for constituent tasks of video panoptic segmentation. The second baseline is based on a network for (image) panoptic segmentation and will add an optical flow based mask propagation and matching module. Depending on what networks are available and what performance is desired, one module can be more relevant for a specific use case. Both modules, however, can be utilized without the need for retraining.

\subsection{Baseline 1: Heuristically merging semantic segmentation and multi-object tracking outputs}

As a first step towards building an algorithm for a complex task, one might first question if there is already something available that can be used. For the first baseline of a model for VPS, two existing models will be used, namely, a model designed and trained for semantic segmentation, i.e. the pixel-wise classification of $stuff$ and $thing$ classes, and a model designed and trained for multi-object tracking i.e. the bounding-box tracking of $thing$ classes through video frames. These two models together give enough information for video panoptic segmentation. The chosen models for the segmentation and tracking are, MobileNetV3-Small \cite{howard2019searching} trained on CityScapes and Deep-SORT \cite{wojke2017simple} trained on the MOT17 dataset, respectively. It must be noted that the multi-object tracking model chosen, can only detect and track pedestrians and not other $thing$ classes. This leads to a video panoptic segmentation format in which only pedestrians are considered to be $things$.


\textbf{Fill \& Fuse}\hspace{0.2cm}
In order to bring the output of the two separate neural networks into the VPS format, the outputs of the segmentation network and of the tracking network will be passed to the Fill \& Fuse module. In this module, these outputs will be heuristically merged, every pixel in a frame sequence is given two scalars, a class label, and an instance ID.
To do this, first the bit-wise AND will be taken of the masks of the segmentation map belonging to the person class, and the whole content of the bounding boxes obtained from the tracking output. This results in a format compatible with video instance segmentation,  \cite{DBLP:journals/corr/abs-1905-04804}, i.e. the video counterpart of (image) instance segmentation \cite{gupta2014learning}. This intermediate result can be seen within the Fill \& Fuse module in Fig. \ref{fig: model1}. Then to bring this format into the video panoptic segmentation format, all pixels that have not been assigned a label will receive their class label directly from the segmentation map and their instance ID will be set to 0.


\subsection{Baseline 2: Mask propagation for time-consistent panoptic segmentation}
\begin{figure*}
\center
\includegraphics[width=\textwidth]{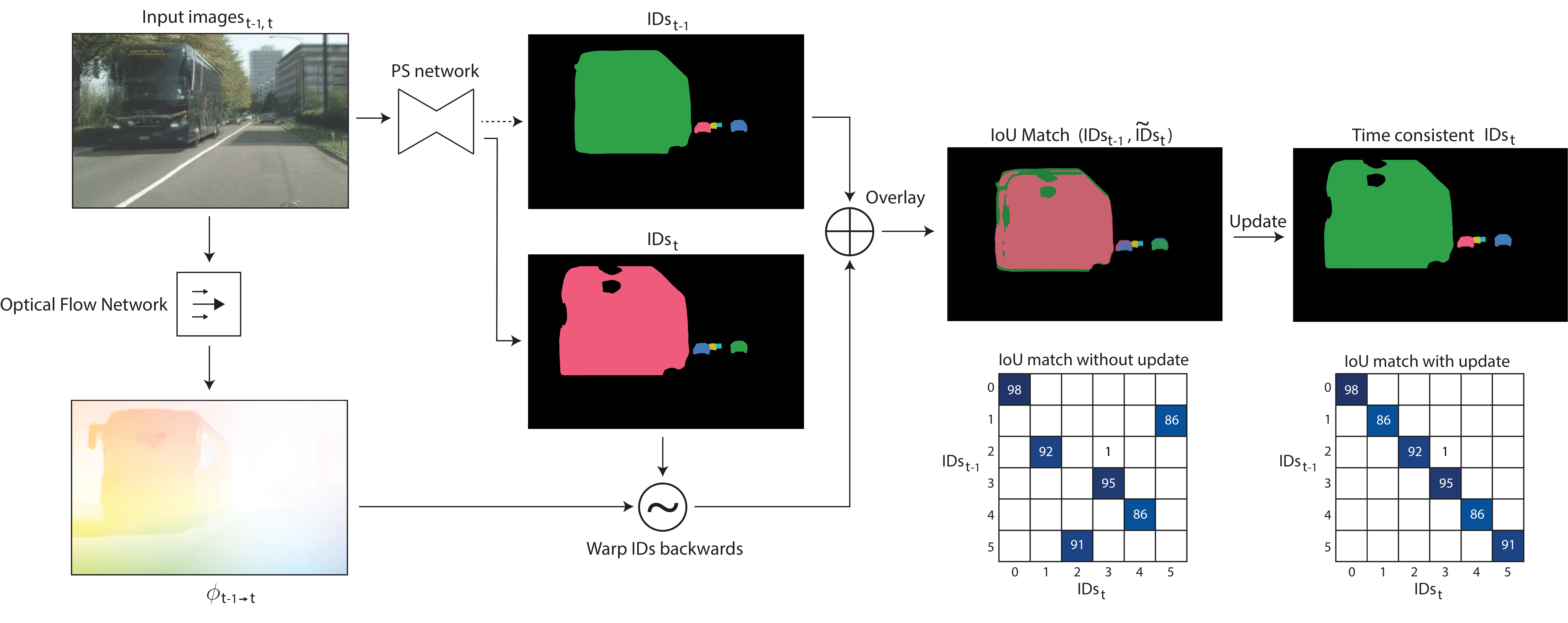}
\caption{Mask warp \& match algorithm. Instance ID masks will get warped backwards using optical flow. This will allow IoU matching between current and previous masks, which can be used to update the time-inconsistent IDs. The confusion matrices show the IoUs (in percentages) between all masks at $t$ and $t-1$, for 6 instance IDs (0-5). The class label channel is left unaltered and is therefore not represented in this figure.}
\label{fig: warping}
\end{figure*}

The proposed module that will be used in our second baseline can be added on top of a network designed for (image) panoptic segmentation in order to add temporal information.
In this work, we will conduct our experiments with a part of VPSNet \cite{Kim2020}. This part consists of an image panoptic segmentation network and an optical flow network and is referred to as VPSNET-Fuse. It is important to point out that this model does not carry any notion of time to track objects as the Track head is removed. VPSNet-Fuse uses a shared backbone followed by two task-specific heads designed for semantic segmentation and instance segmentation. A distinctive feature in this network is their $Fuse$ module, which resides in between their backbone and task heads. This module fuses the extracted features from the current frame and extracted features from an earlier reference frame. The extracted features from a reference frame first get warped using the optical flow \cite{ilg2016flownet}between the frames such that their features have more overlap. It is argued that this module aids in fusing temporal information at the pixel level. For more details about VPSNet, please refer to \cite{Kim2020}.

In this work, we investigate the effects of further utilizing the temporal consistency that dynamic scenes yield, by contributing a module that relies on mask propagation with optical flow to track objects. Therefore, we will not be using optical flow only in between the backbone and the task heads like VPSNet-Fuse, but also apply it to propagate instance masks and explicitly match them to update time-inconsistent instance IDs.

\textbf{Warp \& Match}\hspace{0.2cm}
Since VPS requires only the instance IDs of $thing$ classes to be consistent over time, $stuff$ classes will be ignored while discussing the mask propagation. The proposed tracking method in Fig. \ref{fig: warping} works as follows. First, the input images will be fed through a panoptic segmentation network, yielding $IDs_t$. Note that, the output of this network does not need to be time consistent. In parallel, the optical flow map $\phi_{t-1 \rightarrow t}$, i.e. the pixel-wise displacement, will be estimated, using the input images at time $t$ and $t-1$, by the method proposed in \cite{DBLP:journals/corr/abs-2003-12039}. Secondly, using the inverse flow map $\phi_{t \rightarrow t-1}$, the instance masks at time $t$ can be warped backward, constructing $\widetilde{ID}s_t$, which will approximate the instance masks of the panoptic segmentation network its output one time step earlier i.e. $IDs_{t-1}$. Then, the backward warped instance masks $\widetilde{ID}s_t$ and the instance masks at $t-1$, $IDs_{t-1}$, can be compared using intersection over union (IoU). Lastly, based on the IoU, current masks can be matched with earlier masks and updated if necessary, yielding time-consistent instance IDs masks. The IDs in the first frame, i.e. at $t=0$, do not have to be warped back nor updated. Note that, the class label channel of the network output does not need to be updated. The confusion matrices in Fig. \ref{fig: warping} represent the IoU matching between instance masks before and after an update. In the ideal case, the diagonal of the matrix contains the highest values.


\begin{figure*}
\begin{subfigure}{2\columnwidth}
        \includegraphics[width=\textwidth]{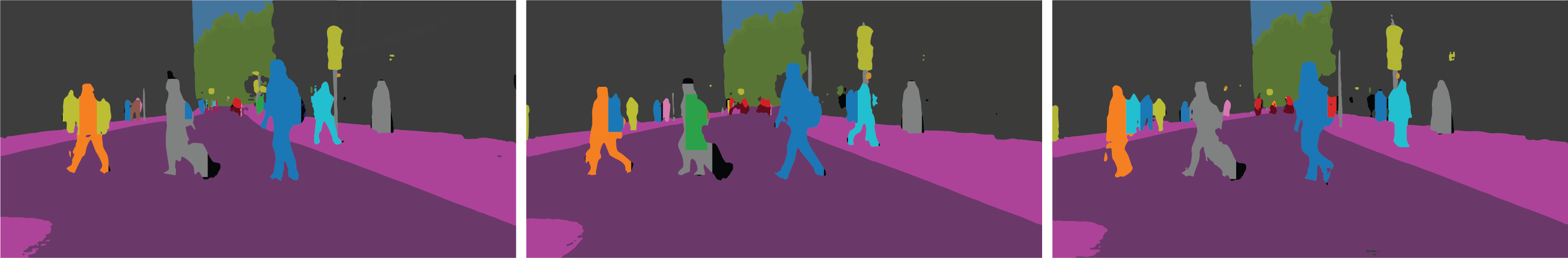}
        \caption{Qualitative results of the first baseline, which uses our Fill \& Fuse module to merge the outputs of semantic segmentation and multi-object tracking.}
        \label{fig: result 1a}
    \end{subfigure}

    \begin{subfigure}{2\columnwidth}
        \includegraphics[width=\textwidth]{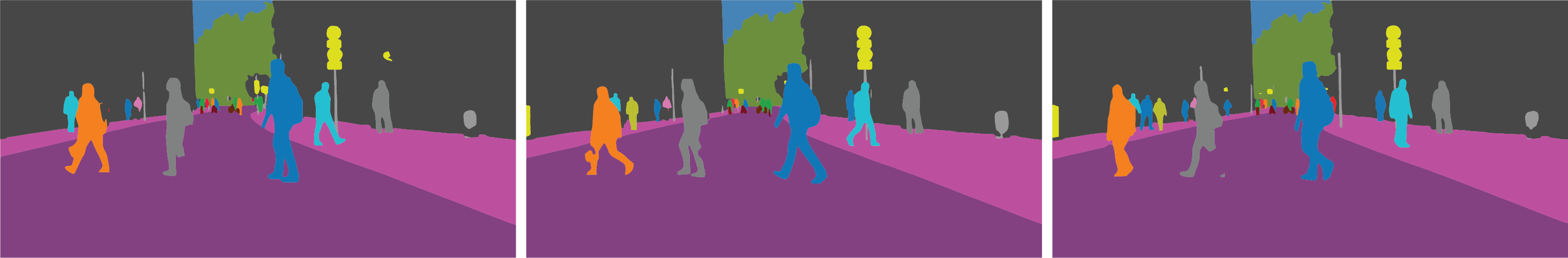}
        \caption{Qualitative results of the second baseline, which uses our Warp \& Match module to bring time-consistent instance IDs to a panoptic segmentation output.}
        \label{fig: result 1b}
    \end{subfigure}

\label{fig: result1}
\caption{Visualized video panoptic segmentation results of both the proposed baselines on the same sequence of the CityScapes dataset. }
\end{figure*}

\section{Discussion and Results}\label{ch: results}

A visualization of the output of the first baseline can be seen in Fig \ref{fig: result 1a}. Because this method only assumed pedestrians as belonging to $thing$ classes, we can only look at the qualitative results and not at the quantitative results an evaluation metric would bring us. From the figure it can be noted that all pixels are given class labels. Vegetation is visualized by the color green, cars by the color blue, and pavement by purple. Since cars were not considered to be countable by this algorithm, all cars have the same color blue. Pedestrians, on the other hand, were considered to be countable and every detected pedestrian is thus given its own instance color. Missed pedestrians however are all red. The crucial factor in the task of video panoptic segmentation is to have the instance IDs (visualized by the different colors for objects with the same class) be consistent over time. Indeed in the visualized sequence of Fig. \ref{fig: result 1a} it can be seen that most pedestrians carry their color across multiple frames.\\
From the same figure, the flaws of the proposed Fill \& Fuse module can also be noted. Within the module, it was assumed that the bounding boxes of pedestrians do not overlap. In case they do overlap, it is not clear how to fuse anymore. This resulted in some rectangular artifacts in the instance masks which especially come forward in the second frame.

The results of the second baseline are shown in Fig. \ref{fig: result 1b}. By visual inspection, it can be noted that the semantics of this model are better than those of the first baseline. Object boundaries are less coarse and there are no rectangular artefacts, as opposed to those of the first baseline. This can be explained by the learnable and joint end-to-end training that has been performed on this model as opposed to the two separate networks in the first baseline. Also, objects at smaller scales are better detected, this results most likely from the pyramidal backbone of the model. The most relevant point that can be observed from this sequence, is that the Warp \& Match module can indeed aid in bringing time-consistent instance Ids. This can be observed from the consistent colors across frames that most objects display. \\
\textbf{Future Work}
We acknowledge that both our proposed modules have still room for improvement and that further investigation is needed to confirm their strength by evaluating them using performance metrics. Regarding the Fill \& Fuse module, future research can include the examining of how to handle fusing instance masks when bounding boxes overlap. Regarding the Warp \& Match module, up to now, this module only updates instance ID masks that are present in consecutive frames, but ignores ID masks that vanish in one frame and return in a later one. This is not optimal in the case where a certain object has been detected as an instance in one frame, but is only given a class label in the next frame. Solving this issue makes the module also applicable for more dynamic scenes where objects are not always detected.

\section{Conclusion}\label{ch: conclusion}

We propose two modules that can be used to augment the functionality of existing models for scene understanding. The first one is our fuse \& fill module, which merges the outputs of semantic segmentation and multi-object tracking to create video panoptic segmentation. The second one is our warp \& match module, which extends panoptic segmentation to the time-domain using optical flow, also creating video panoptic segmentation. Both modules rely on heuristics and can thus be added to any model that is in the required format, without the need for retraining. The potential of both modules has been shown by the qualitative results presented.

\bibliography{bibliography.bib}

\end{document}